\documentclass[journal]{IEEEtran}
\pdfoutput=1
\usepackage{amsmath} 
\usepackage[english]{babel}

\usepackage{graphicx} 
\usepackage{enumerate}
\usepackage{graphics}
\usepackage{enumitem}
\usepackage{amssymb}
\usepackage{float}
\usepackage{rotating}
\usepackage{comment}
\usepackage{copyrightbox}
\usepackage{rotating}
\usepackage{booktabs}
\usepackage{colortbl}
\usepackage{multicol}
\usepackage{multirow}

\usepackage{cite}
\usepackage{hyperref}

\newcommand{\proba}{\text{p}}

\newcommand{\V}[1]{\boldsymbol{#1}}
\newcommand{\M}[1]{\mathbf{#1}}
\newcommand{\transpose}{^\top}

\newcommand{\FT}[1]{\introducedinrevision{#1}}

\newcommand{\introducedinrevision}[1]{\textcolor{black}{#1}}

\begin{document}
\markboth{\centering IEEE TRANSACTIONS ON GEOSCIENCE AND REMOTE SENSING, PREPRINT. PUBLISHED VERSION (2021): 10.1109/TGRS.2021.3128621}{}

\title{As if by magic: self-supervised training\\ of deep despeckling networks with MERLIN}

\author{Emanuele~Dalsasso, Lo{\"i}c~Denis, Florence~Tupin%
\thanks{E. Dalsasso and F. Tupin are with LTCI, Télécom Paris, Institut Polytechnique de Paris, Palaiseau, France, e-mail: forename.name@telecom-paris.fr.}%
\thanks{L. Denis is the Univ Lyon, UJM-Saint-Etienne, CNRS, Institut d Optique Graduate School, Laboratoire Hubert Curien UMR 5516, F-42023, SAINT-ETIENNE, France, e-mail: loic.denis@univ-st-etienne.fr.}%
\vspace{-1\baselineskip}}

\IEEEaftertitletext{\centering This is the pre-acceptance version, to read the final version published in 2021 in the IEEE Transactions on Geoscience and Remote Sensing (IEEE TGRS), please go to: \textcolor{blue}{10.1109/TGRS.2021.3128621}\vspace{1\baselineskip} }

\maketitle

\begin{abstract}
Speckle fluctuations seriously limit the interpretability of synthetic aperture radar (SAR) images. Speckle reduction has thus been the subject of numerous works spanning at least four decades. Techniques based on deep neural networks have recently achieved a new level of performance in terms of SAR image restoration quality.

Beyond the design of suitable network architectures or the selection of adequate loss functions, the construction of training sets is of uttermost importance. So far, most approaches have considered a supervised training strategy: the networks are trained to produce outputs as close as possible to speckle-free reference images. Speckle-free images are generally not available, which requires resorting to 
\FT{natural or optical images}
or the selection of stable areas in long time series to circumvent the lack of ground truth. Self-supervision, on the other hand, avoids the use of speckle-free images.

We introduce a self-supervised strategy based on the separation of the real and imaginary parts of single-look complex SAR images, called MERLIN (coMplex sElf-supeRvised despeckLINg), and show that it offers a straightforward way to train all kinds of deep despeckling networks. Networks trained with MERLIN take into account the spatial correlations due to the SAR transfer function specific to a given sensor and imaging mode. By requiring only a single image, and possibly exploiting large archives, MERLIN opens the door to hassle-free as well as large-scale training of despeckling networks. The code of the trained models is made freely available at \url{https://gitlab.telecom-paris.fr/RING/MERLIN}.


\end{abstract}
\begin{IEEEkeywords}
SAR, image despeckling, deep learning, self-supervised training.
\end{IEEEkeywords}

\IEEEpeerreviewmaketitle

\section{Introduction}
\IEEEPARstart{T}{he} speckle phenomenon occurs due to the coherent summation of many elementary echoes within a radar resolution cell. It is responsible for the strong fluctuations that dramatically degrade the quality of SAR images. Speckle suppression has been the subject of many research works, from the pioneering works of Lee \cite{lee1983digital} to the most recent techniques based on deep neural networks \cite{fracastoro2020deep,zhu2021deep,denis2021review}.

The first approaches developed to reduce the speckle fluctuations were based on a local averaging of the pixel intensities within a small window. To prevent the spreading of bright targets over the whole window, selection techniques were introduced to exclude pixels with values too different from the value at the reference pixel
\cite{lee1983digital,vasile2006intensity}, or to select an oriented window with a homogeneous content \cite{lopes1990adaptive}. Regularization techniques formulate the restoration problem as the minimization of the sum of a data fidelity term, favoring restored images statistically close to the speckled one, and a regularization term that encourages the solution to be smooth. To prevent the apparition of a blur, edge-preserving terms such as the total-variation have been widely applied \cite{aubert2008variational,denis2009sar,bioucas2010multiplicative,steidl2010removing}. To preserve both point-like structures and smooth areas with sharp boundaries, image decomposition models were introduced \cite{aujol2003image,lobry2016multitemporal}. Wavelets \cite{xie2002sar} and curvelets \cite{durand2009multiplicative} were also applied in several works.

The large success of patch-based methods in image denoising \cite{NLM,BM3Dwiener} fueled a large body of research \cite{deledalle2014exploiting,tupin2019ten}, from intensity-image restoration \cite{deledalle2009iterative,parrilli2011nonlocal,cozzolino2013fast} to interferometric \cite{deledalle2010nl}, polarimetric \cite{chen2010nonlocal,torres2014speckle} or polarimetric and interferometric \cite{deledalle2014nl} modalities.

There has been a revival of interest in the restoration of SAR intensity images with the advent of deep learning. The most distinctive feature of deep despeckling networks is the training approach \cite{denis2021review}: (i) the first generation of methods used supervised training techniques where pairs of speckled/speckle-free images were formed to train the network; (ii) then, self-supervised techniques used pairs of images of the same area, acquired at different times; (iii) self-supervised training with a single image represents the ultimate goal.

A fundamental limitation of supervised training (i) is the difficulty to obtain the speckle-free image associated with a speckled image. Most techniques from this category, therefore, rely on synthetic speckle, i.e., the simulation of speckle corruption, starting from an optical image \cite{wang2017sar,wang2017generative,zhang2018learning} or from the temporal mean of a long time series of SAR images \cite{lattari2019deep,dalsasso2020sar}. In the speckle simulations, however, spatial correlations are typically ignored, which requires that actual SAR images be pre-processed before applying the network (by resampling and inversion of the SAR transfer function \cite{lapini2013blind,abergel2018subpixellic}, or down-sampling \cite{dalsasso2020handle}). A way to circumvent this problem is to use actual SAR images as input and a temporal average as reference \cite{chierchia2017sar,cozzolino2020nonlocal}. Any change that occurs during the time series is a potential source for bias. Temporally stable areas must then be selected, which is challenging, in particular with cultivated lands, since a trade-off between sufficient speckle reduction (hence, large temporal series) and limited changes must be found.

Self-supervised training (ii) uses pairs formed by two images of the same area, acquired at separate dates so that the speckle is temporally decorrelated. That way, the second image, though noisy, can be used as a reference for the network. For this to work in practice, changes between the two images must be compensated \cite{dalsasso2021sar2sar}, which requires in turn resorting to a despeckling technique. Compared to supervised techniques, this approach benefits from training with actual SAR images and it is thus robust to spatial correlations of speckle due to the SAR transfer function.

Finally, self-supervised training based on a single-image (iii) uses networks with a specific architecture \cite{9324183,molini2020speckle2void} so that the receptive field does not include the central area, forming a blind-spot \cite{laine2019high} that can be used to supervise the estimation by the network. The very specific receptive field of such networks strongly constrains their architecture, which can limit their performance. Moreover, for the self-supervision to succeed, the speckle must be spatially decorrelated, which requires the same pre-processing techniques as described previously. Several variants of the concepts of blind-spot have been recently developed in the literature of image denoising, using various masking strategies \cite{lee2020noise2kernel,xie2020noise2same}. The spatial correlation of the speckle represents, however, a severe limitation to the potential application of these approaches to SAR imaging.

\medskip
\emph{Our contributions:}
This paper introduces a new training strategy, applicable to all kinds of network architectures. This strategy is fully unsupervised: it only requires single-look complex (SLC) images to perform the training or to process new data once the network is trained. In contrast to other existing works, it does not require additional hypotheses like the absence of spatial correlations of the speckle, or temporal stability throughout a time series. The phase information of single-look complex images is often considered irrelevant when only the intensity is of interest (i.e., apart from the context of SAR interferometry). Finding an interest in the real and imaginary components for the restoration of intensity images may even seem disconcerting. In section \ref{sec:statmod}, we derive a statistical model of SLC images showing that two independent and identically distributed images can be extracted from an SLC image. This paves the way to the application of a self-supervised training strategy inspired by noise2noise \cite{lehtinen2018noise2noise}: MERLIN, see section \ref{sec:MERLINdescr}. Results obtained in section \ref{sec:results} on images at different spatial resolutions confirm the ability of MERLIN to produce high-quality restoration results at medium, high, and very high spatial resolutions.

\section{Statistical model of SAR images}
\label{sec:statmod}

\begin{figure*}[!t]
    \centering
    \includegraphics[width=\textwidth]{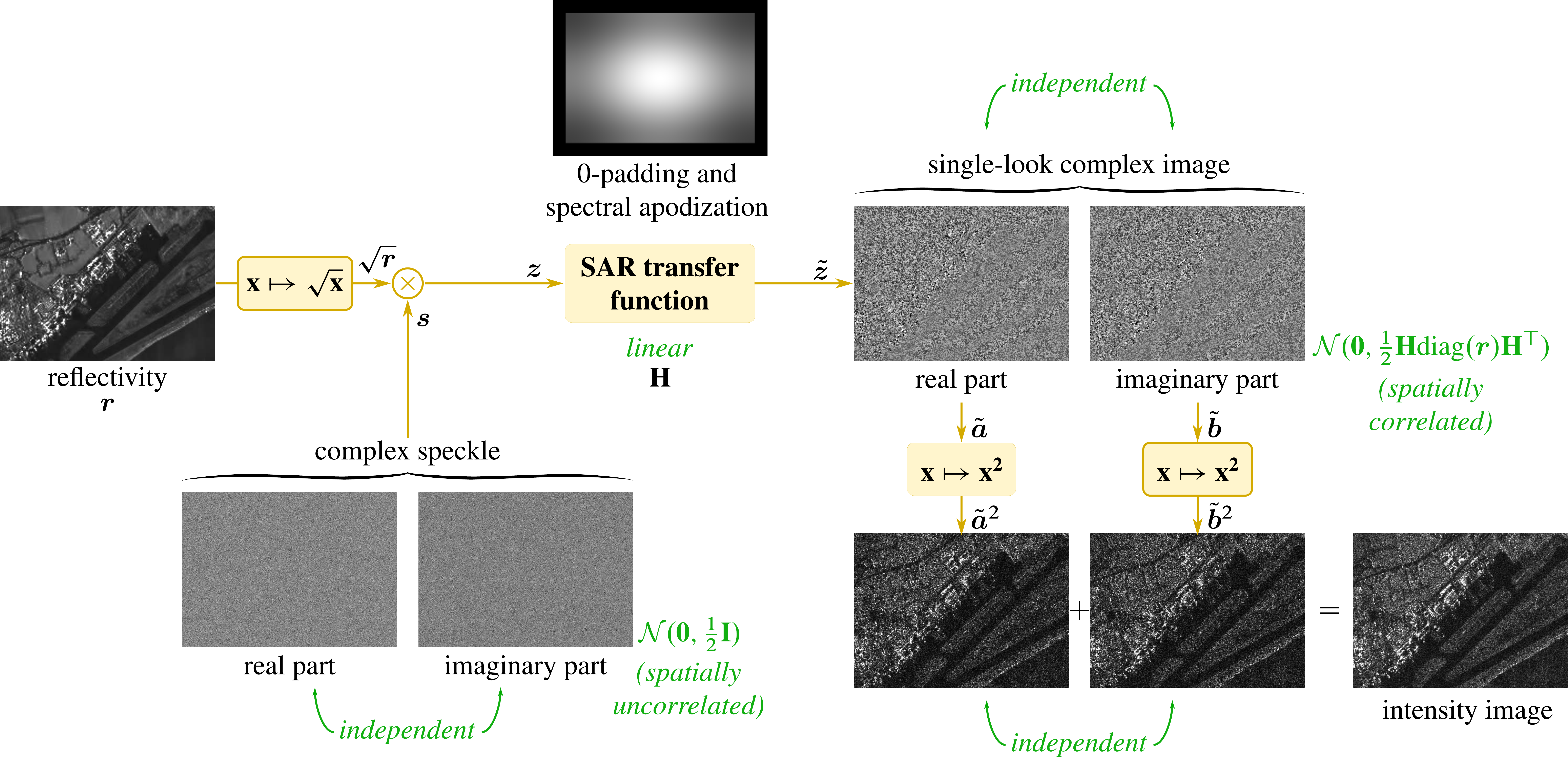}
    \caption{A statistical model of speckle in SAR image: the intensity image on the right is a corrupted version of the reflectivity image shown on the left. The single-look complex image contains spatially-correlated speckle components that are independent in the real and imaginary parts. The SAR transfer function shown here corresponds to Sentinel-1 stripmap mode. For visualization purposes, a non-linear look-up table is used to display intensity images.}
    \label{fig:statsSAR}
\end{figure*}

Goodman's fully developed speckle model \cite{goodman2007speckle} gives the statistical distribution followed by the complex amplitude $z=\introducedinrevision{\rho}\exp(j\varphi)$ resulting from the coherent summation of many elementary echoes $\introducedinrevision{\rho}_n\exp(j\varphi_n)$. Under the hypothesis of a large surface roughness compared to the radar wavelength, each elementary echo has a phase that is independent and uniformly distributed in the range $[-\pi,\pi]$. If the area in the radar resolution cell is homogeneous, elementary amplitudes $\introducedinrevision{\rho}_n>0$ are also independent and identically distributed (i.i.d.) as well as independent from the phases $\varphi_n$. It follows from the central limit theorem, in the limit of a large number of elementary echoes, that the distribution of $z=\sum_n \introducedinrevision{\rho}_n\exp(j\varphi_n)$ converges to a circular complex Gaussian distribution \cite{goodman2007speckle}:
\begin{align}
    \proba_Z(z) = \frac{1}{\pi r}\exp(-|z|^2/r)\,,
\end{align}
where $z$ is the complex amplitude at a given pixel and $r>0$ is the SAR reflectivity at that pixel.

The multiplicative nature of speckle phenomenon becomes clear by writing $z$ under the form $z=s\sqrt{r}$, with $r$ the reflectivity of the homogeneous area and $s$ a complex random variable distributed according to:
\begin{align}
    \proba_S(s) = \frac{1}{\pi}\exp(-|s|^2)\,.
\end{align}
The decomposition of $z$ into its real and imaginary parts, $z=a+jb$, leads to:
\begin{align}
    \proba_Z(z) &= \proba_Z(a+jb) =\frac{1}{\pi r}\exp(-(a^2+b^2)/r)\nonumber\\
    &=\underbrace{\frac{1}{\sqrt{2\pi}\sqrt{r/2}}\exp(-a^2/r)}_{\mathcal{N}(0,r/2)}\underbrace{\frac{1}{\sqrt{2\pi}\sqrt{r/2}}\exp(-b^2/r)}_{\mathcal{N}(0,r/2)}\,,
\end{align}
which shows that the real and imaginary parts of the complex amplitude are i.i.d. according to a Gaussian distribution with variance $r/2$, or equivalently, that the real and imaginary parts of the complex-speckle component $s$ in the multiplicative model are i.i.d. and Gaussian-distributed with a variance equal to 1/2.

This multiplicative stochastic model is illustrated in the left part of figure \ref{fig:statsSAR}. From one pixel to the next, the realization of speckle is different and the random field $\V s\in\mathbb{C}^K$ of a $K$-pixels image is a white Gaussian field.

Depending on the acquisition mode, the chosen pixel size, and the spectral apodization applied to reduce sidelobes around bright targets, a specific SAR transfer function then transforms the spatially uncorrelated field $\V z $ into a spatially correlated field $\tilde{\V z}$, see Fig.\ref{fig:statsSAR} (center):
\begin{align}
    \tilde{\V z} &= \M H \V z\,,
\end{align}
with $\M H$ the spatial-domain operator associated to the SAR transfer function.
By linearity of $\M H$, we get:
\begin{align}
    \tilde{\V a} &= \M H \V a \quad\text{ and }\quad\tilde{\V b} = \M H \V b\,.
\end{align}
\introducedinrevision{If the SAR system $\M H$ is real-valued (for a shift-invariant system, this corresponds to a frequency response with Hermitian symmetry), then 
$\tilde{\V a}$ and $\tilde{\V b}$ are spatially correlated but \emph{mutually independent} random fields (appendix \ref{sec:appendixindep} derives 
a slightly more general condition
on $\M H$ to obtain statistically independent components $\tilde{\V a}$ and $\tilde{\V b}$).} The element-wise multiplication by $\sqrt{\V r}$ and the linear operation $\M H$ transform the white Gaussian fields corresponding to the real and imaginary parts of $\V s$, distributed according to $\mathcal{N}(\V 0,\frac{1}{2}\M I)$, into two i.i.d. Gaussian fields $\tilde{\V a}$ and $\tilde{\V b}$ distributed according to $\mathcal{N}(\V 0,\frac{1}{2}\M H\text{diag}(\V r)\M H\transpose)$, with $\text{diag}(\V r)$ the $K\times K$ diagonal matrix whose diagonal is equal to the vector $\V r\in\mathbb{R}_{+*}^K$.

Finally, the intensity image in SAR imaging is obtained by summing the squared real and imaginary parts, see the bottom right of Fig.\ref{fig:statsSAR}. The square is applied separately to the real and imaginary components. Random fields $\tilde{\V a}^2$ and $\tilde{\V b}^2$ are thus still i.i.d.

In summary, as depicted in Fig.\ref{fig:statsSAR}, the statistical model of SAR image formation shows that the two components $\tilde{\V a}^2$ and $\tilde{\V b}^2$, corresponding to the squared real and imaginary part of an SLC image that are added to form the SAR intensity image, are \emph{independent and identically distributed}. Each component contains half of the information, or, in other words, has a signal-to-noise ratio (SNR) that is $1/\sqrt{2}$ times the SNR of the intensity image (the intensity corresponds to the sum of these two independent components, its variance is halved, which corresponds to a SNR improvement by a factor $\sqrt{2}$).

\section{MERLIN: complex self-supervised despeckling}
\label{sec:MERLINdescr}

\begin{figure*}[!t]
    \centering
    \includegraphics[width=\textwidth]{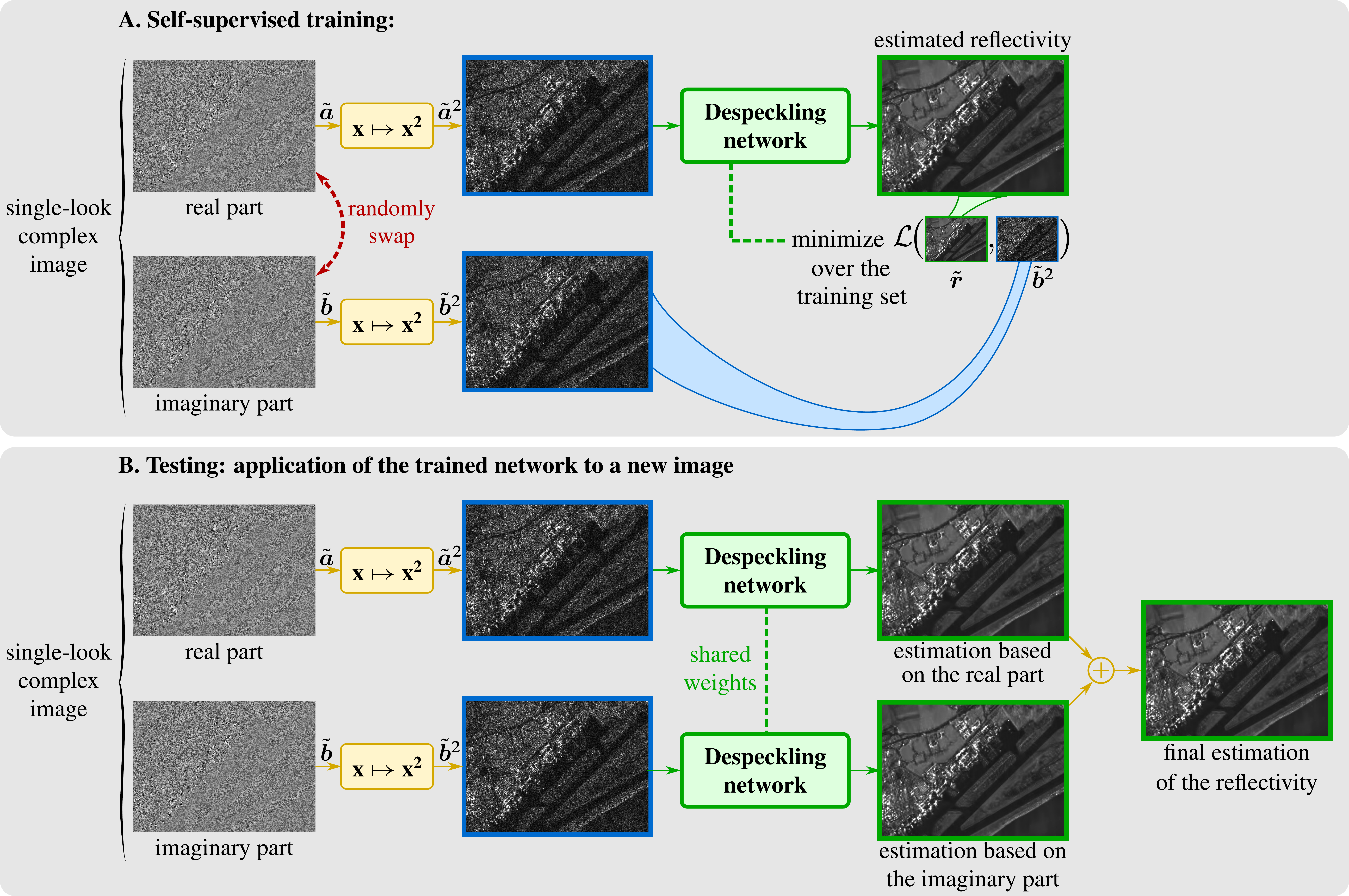}
    \caption{The principle of MERLIN: during step A, the despeckling network is trained to estimate the reflectivity based solely on the real part. The loss function evaluates the likelihood of the predictions according to the imaginary part. Once the network is trained, it can be used as shown in B: the real and imaginary parts are processed separately using networks with the same weights. The outputs are combined to form the final estimation. Note that, to simplify the figure, step A is illustrated only with the real part as input but real and imaginary parts can be swapped during training to increase the number of training samples, see text.}
    \label{fig:MERLINppl}
\end{figure*}

Since SLC images provide two i.i.d. components that contain half the information\footnote{we consider here that the raw phase $\text{angle}(\V z)$ is non-informative, which of course is no longer true when considering multiple SLC images in interferometric configuration} from the intensity image, a self-supervised training strategy can be built by processing one component (e.g., the real part) and evaluating the restoration quality on the other component (e.g., the imaginary part). This corresponds to an ideal application case of the noise2noise principle \cite{lehtinen2018noise2noise} in which a deep neural network is trained to predict a noisy image from another independent noisy realization. Since the realization-specific random perturbation can not be guessed by the network, it tends to remove the noise in the input image even if no noiseless image is provided to the loss function.

We follow a similar approach to train networks by performing a coMplex sElf-supeRvised despeckLINg (MERLIN). Our approach is graphically summarized in figure \ref{fig:MERLINppl}: during the training phase (step A of the figure), the network is trained to process only the real part and a loss function is evaluated to measure how close the estimated reflectivity is to the imaginary component. Once the network is trained, it can be applied to reduce speckle noise in SLC images (step B of the figure). This time, both the real and imaginary parts are independently processed using the same network weights (i.e., a single network is trained in step A). The two estimations are then combined to produce the final estimation (by averaging). When despeckling SLC images, all the information is used, i.e., both the real and imaginary parts.

In order to define the loss function used during the training phase (step A), it is necessary to decide which parameters should be estimated. The SAR transfer function has a significant impact on the image appearance: the 0-padding controls the pixel size while the spectral apodization sets the height of the sidelobes. Rather than inverting the SAR transfer function, we consider producing an image with the same characteristics (pixel size and bright point signature). With an ideal SAR transfer function $\M H=\M I$ (the identity matrix of dimension $K\times K$), the real and imaginary parts $\tilde{\V a}$ and $\tilde{\V b}$ have a variance equal to $\V r/2$. With a non-ideal transfer function, the variance corresponds to the diagonal of matrix $\frac{1}{2}\M H\text{diag}(\V r)\M H\transpose$, i.e., the variance of the $k$-th pixel is $\tilde{r}_k/2$ with $\tilde{r}_k=\sum_\ell\text{H}_{k\ell}^2r_\ell$. With MERLIN, we aim at estimating the values $\tilde{r}_k$ for each pixel. The marginal distribution of $\tilde{a}_k$ and $\tilde{b}_k$ (the real and imaginary parts at pixel $k$ of the SLC image) is a centered Gaussian with variance $\tilde{r}_k/2$. We thus define the following loss function $\mathcal{L}$:
\begin{align}
    \mathcal{L}(\tilde{\V r},\tilde{\V b})=
    \sum_k \frac{1}{2}\log\left(\tilde{r}_k\right)+\frac{\tilde{b}_k^2}{\tilde{r}_k}\,,
    \label{eq:loss}
\end{align}
which corresponds, up to an additive constant, to the sum over all pixels of the opposite of the log-likelihood of the marginal distribution. To reduce the dynamic range, it is beneficial that the network inputs and outputs be expressed in log-scale. We introduce $\check{r}_k=\log \tilde{r}_k$, $\check{a}_k=\log |\tilde{a}_k|$, and $\check{b}_k=\log |\tilde{b}_k|$ to define the equivalent loss function expressed with log-scale images:
\begin{align}
    \mathcal{L}_{\text{log}}(\check{\V r},\check{\V b})=
    \sum_k \frac{1}{2}\check{r}_k+\exp\!\left(2\check{b}_k-\check{r}_k\right)\,.
    \label{eq:losslog}
\end{align}

Note that since the real and imaginary parts are i.i.d. and given that we use the same network weights in step B to process both the real and imaginary parts, the training phase (step A) can not only be performed with the real part as input and the loss $\mathcal{L}(\tilde{\V r},\tilde{\V b})$, but also with the imaginary part as input and the loss $\mathcal{L}(\tilde{\V r},\tilde{\V a})$ (only the former case is represented in Fig.\ref{fig:MERLINppl} for simplicity reasons while both are applied in practice).

\section{Experimental validation}
\label{sec:results}

\begin{figure*}[!t]
    \centering
    \includegraphics[width=\textwidth]{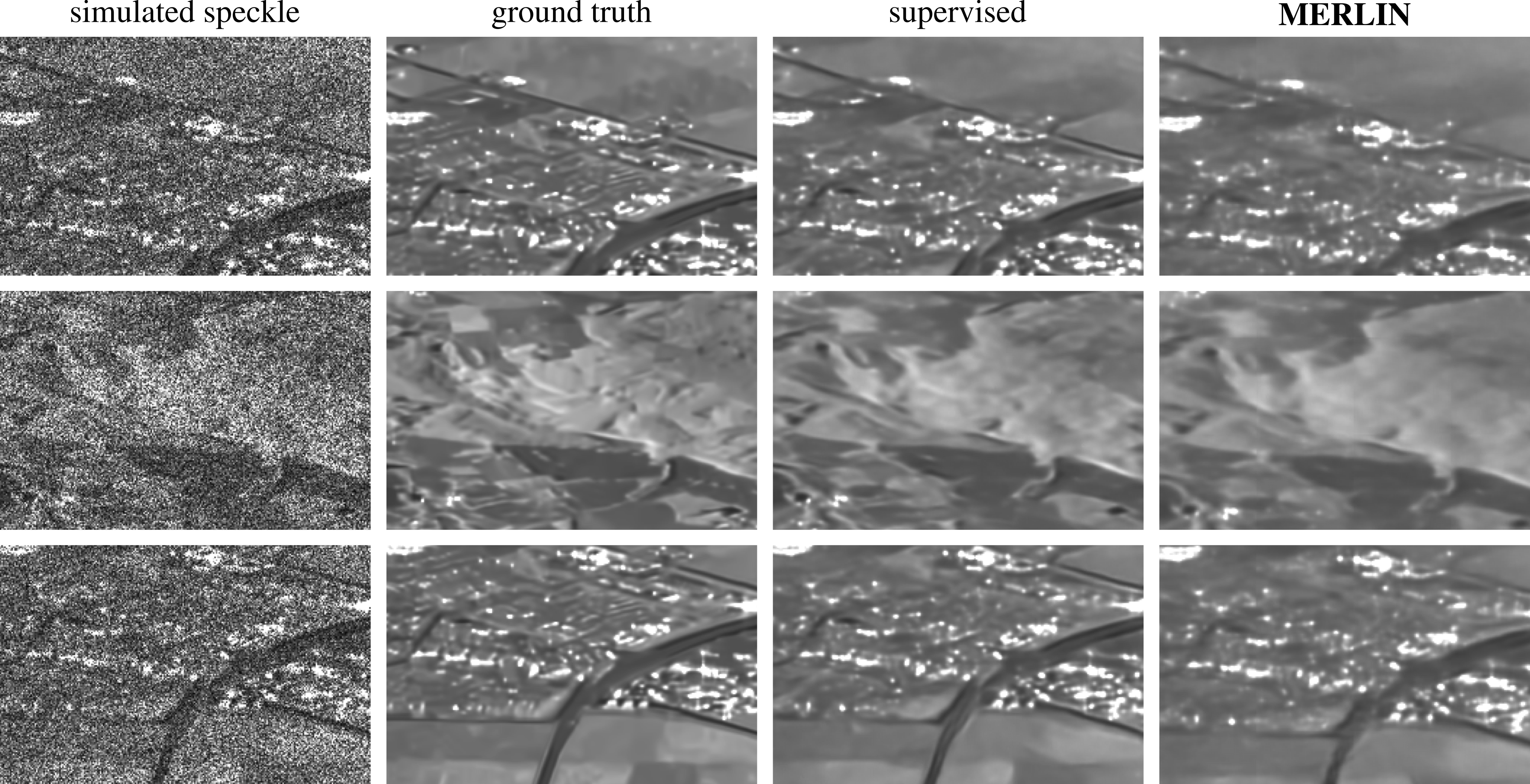}
    \caption{Despeckling results on images corrupted by synthetic speckle: the first column shows the noisy images obtained by multiplying the ground truth images $\sqrt{\V r}$ shown in the second column by a white speckle field. The third column gives despeckling results obtained with a U-Net trained in a supervised fashion (SAR2SAR, step A \cite{dalsasso2021sar2sar}). The last column gives despeckling results obtained with the same network trained with MERLIN. Compared to the supervised training, MERLIN is penalized because it only has access to a single speckle realization \FT{through the real or imaginary part}.
    The images shown in each row are regions of interest extracted from the images Lely, Limagne, and Marais1 that appear in table \ref{table:comparison_psnr}. Additional images can be seen on
    \url{https://gitlab.telecom-paris.fr/RING/MERLIN}.
    }
    \label{fig:resultsMERLINsimus}
\end{figure*}

In contrast to the family of self-supervised methods derived from the concept of blind-spot \cite{laine2019high} that require the receptive field of the network to exclude the central pixel(s), MERLIN imposes no constraint on the type of neural network used to perform the estimation. In the following experiments, we used a simple U-Net architecture \cite{ronneberger2015u}. This network, originally developed for semantic segmentation, performs very well on image denoising tasks and can be trained quickly \cite{lehtinen2018noise2noise,lattari2019deep,dalsasso2021sar2sar}.




\begin{table*}[htpb]
    \centering
    \caption{Description of the training parameters for all experiments carried out with a residual U-Net trained with MERLIN.}
\resizebox{\textwidth}{!}{
\begin{tabular}{lc@{\hspace*{5ex}} c@{\hspace*{5ex}} c@{\hspace*{5ex}} c}
\toprule
                                                                  & \bf Synthetic speckle                & \bf TerraSAR-X stripmap              & \bf TerraSAR-X \introducedinrevision{HS} spotlight             & \bf SETHI                    \\
\midrule
\bf \# images                                                         & 7                         & 3                         & 4                         & 1                          \\ 
\bf patch size                                                        & $256\times 256$           & $256\times 256$           & $256\times 256$           & $256\times 256$           \\ 
\bf batch size                                                        & 12                        & 12                        & 12                        & 12                         \\ 
\introducedinrevision{\bf \# patches}                                                        & \introducedinrevision{12420}                      & \introducedinrevision{50604}                      & \introducedinrevision{27048}                      & \introducedinrevision{76260}                     \\ 
\bf \# batches                                                        & 1035                      & 4217                      & 2254                      & 6355                      \\ 
\bf \# epochs                                                         & 30                        & 30                        & 30                        & 30                        \\ 
\bf gradient norm \cite{zhang2019gradient} & 1.0                       & 1.0                       & 0.5                       & 1.0                       \\ 
\multirow{3}{*}{\bf learning rate $\Biggl\{$}                                    & \introducedinrevision{$10^{-3}$}                 & \introducedinrevision{$10^{-3}$}                 & \introducedinrevision{$10^{-3}$}                 & \introducedinrevision{$10^{-3}$}                 \\ 
                                                                  & \introducedinrevision{$10^{-4}$} after 6 epochs  & \introducedinrevision{$10^{-4}$} after 4 epochs  & \introducedinrevision{$10^{-4}$} after 4 epochs  & \introducedinrevision{$10^{-4}$} after 3 epochs  \\ 
                                                                  & \introducedinrevision{$10^{-5}$} after 20 epochs & \introducedinrevision{$10^{-5}$} after 20 epochs & \introducedinrevision{$10^{-5}$} after 20 epochs & \introducedinrevision{$10^{-5}$} after 20 epochs \\ 
                                                                  \bottomrule
\end{tabular}}
\label{table:hyperparameters}
\end{table*}

To limit the dynamic range of the images at the input of the network, images $\tilde{\V a}$ and $\tilde{\V b}$ are log-transformed and normalized using a fixed affine transform.


In the following set of images, a residual U-Net (based on the network described in \cite{lehtinen2018noise2noise}) has been trained on SLC SAR images. The experiments have been conducted both on images with synthetic speckle noise and on real SAR images. The hyper-parameters used to train the network for each imaging modality are listed in table \ref{table:hyperparameters}. The weights of the trained models are made available for testing at \url{https://gitlab.telecom-paris.fr/RING/MERLIN}.

\subsection{Evaluation of MERLIN on images with synthetic speckle}
\label{sec:results_simus}

\begin{table*}[t]
	\centering
	\caption{Comparison of denoising quality in terms of PSNR on
      amplitude images. For each ground truth image, 20 noisy
      instances are generated. 1$\sigma$ confidence intervals are
      given. Per-method averages are indicated at the bottom.}
	\resizebox{\textwidth}{!}{
	\begin{tabular}{l c c@{\hspace*{4ex}} c@{\hspace*{3ex}} c@{\hspace*{2ex}} c @{\hspace*{2ex}}c@{\hspace*{1.5ex}} c@{\hspace*{1ex}} c}
	\toprule
	Images      & Noisy            & SAR-BM3D          & NL-SAR            & MuLoG+BM3D       & SAR-CNN  & SAR2SAR\textsubscript{A} & Speckle2Void & \textbf{MERLIN}\\
	      &             & (patch-based)          & (patch-based)            & (patch-based)       & (deep network)   & (deep network) & (deep network)& (deep network)\\
	      &             &           &             &        & (supervised)   & (supervised) & (self-supervised)& (self-supervised)\\
    \midrule
	Marais 1    & 10.05$\pm$0.014 & 23.56$\pm$0.134 & 21.71$\pm$0.126 & 23.46$\pm$0.079 & 24.65$\pm$0.086 & \textbf{25.73}$\pm$0.125 & 24.89$\pm$0.102 & 25.25$\pm$0.113\\
	Limagne     & 10.87$\pm$0.047 & 21.47$\pm$0.309 & 20.25$\pm$0.196 & 21.47$\pm$0.218 & 22.65$\pm$0.291 & \textbf{24.45}$\pm$0.119 & 23.40$\pm$0.121 & 23.86$\pm$0.111\\
	Saclay      & 15.57$\pm$0.134 & 21.49$\pm$0.368 & 20.40$\pm$0.270 & 21.67$\pm$0.244 & 23.47$\pm$0.228 & \textbf{23.60}$\pm$0.437 & 19.00$\pm$0.481 & 22.34$\pm$0.484\\
	Lely        & 11.45$\pm$0.005 & 21.66$\pm$0.445 & 20.54$\pm$0.330 & 22.25$\pm$0.437 & \textbf{23.79}$\pm$0.491 & 23.67$\pm$0.542 & 19.28$\pm$0.575 & 22.85$\pm$0.467 \\
	Rambouillet &  8.81$\pm$0.069 & 23.78$\pm$0.198 & 22.28$\pm$0.113 & 23.88$\pm$0.169 & \textbf{24.73}$\pm$0.080 & 24.16$\pm$0.385 & 21.47$\pm$0.488 & 23.53$\pm$0.316\\
	Risoul      & 17.59$\pm$0.036 & 29.98$\pm$0.264 & 28.69$\pm$0.201 & 30.99$\pm$0.376 & \textbf{31.69}$\pm$0.283 & 30.68$\pm$0.230 & 21.42$\pm$0.119 & 29.80$\pm$0.138\\
	Marais 2    &  9.70$\pm$0.093 & 20.31$\pm$0.783 & 20.07$\pm$0.755 & 21.59$\pm$0.757 & 23.36$\pm$0.807 & \textbf{26.63}$\pm$0.215 & 25.04$\pm$0.222 & 26.10$\pm$0.193\\
	\midrule
	Average    &  12.00 & 23.17 & 21.99 & 23.62 & 24.91 & \textbf{25.56} & 22.07 & 25.13\\
    \bottomrule		
	\end{tabular}}
	\label{table:comparison_psnr}
\end{table*}

\begin{figure*}[!ht]
    \centering
    \includegraphics[width=\textwidth]{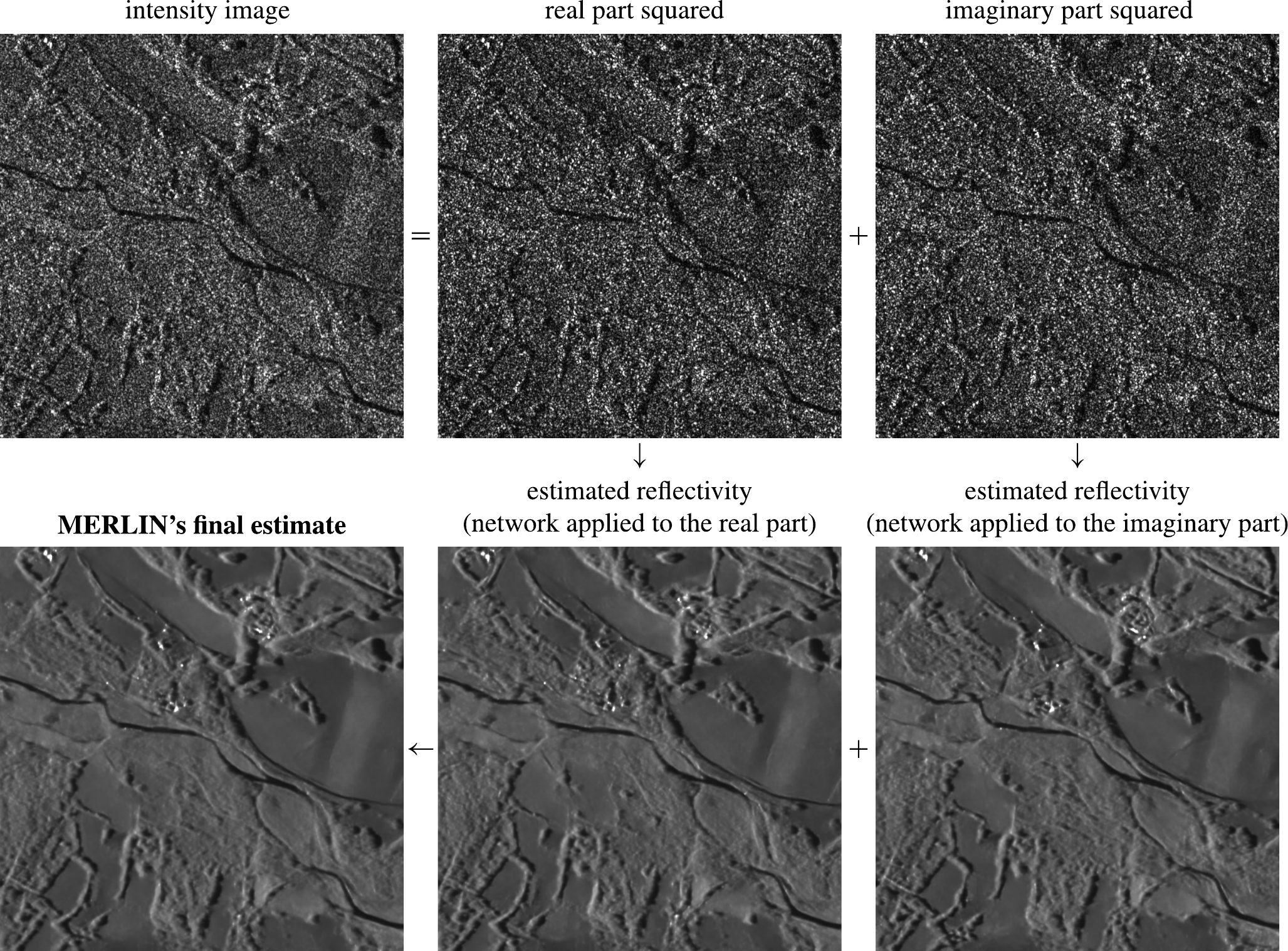}
    \caption{Application of a U-Net trained with MERLIN on a TerraSAR-X image near Serre-Ponçon dam, in the French Alps, acquired in stripmap mode. Additional despeckling results on TerraSAR-X images in stripmap mode can be seen on \url{https://gitlab.telecom-paris.fr/RING/MERLIN}.}
    \label{fig:resultsMERLINreal1}
\end{figure*}

We first evaluate the capability of MERLIN to train a network to restore images synthetically corrupted by speckle. The training set of speckle-free images has been built according to \cite{dalsasso2020sar}. We consider an ideal SAR transfer function: $\M H=\M I$ so that many different despeckling techniques can be applied and compared. Figure \ref{fig:resultsMERLINsimus} compares restoration results on 3 different images with simulated speckle, for the same network architecture but two different training strategies: (i) a supervised training where the network has access to the full intensity image (i.e., $\tilde{\V a}^2+\tilde{\V b}^2$) and the loss function is evaluated on a different, independent, speckle realization drawn from the same ground-truth image (i.e., step A of SAR2SAR algorithm \cite{dalsasso2021sar2sar}, \FT{SAR2SAR$_{\text{A}}$}); (ii) a self-supervised training with MERLIN where the network only has access to either $\tilde{\V a}^2$ or $\tilde{\V b}^2$ and the loss function is $\mathcal{L}(\check{\V r},\check{\V b})$ or $\mathcal{L}(\check{\V r},\check{\V a})$, respectively. Note that in the approach (i) the loss function used during training corresponds to:
\begin{align}
    \mathcal{L}_{\text{log}}(\check{\V r},\check{\V a}')+\mathcal{L}_{\text{log}}(\check{\V r},\check{\V b}')=
    \sum_k \check{r}_k+\exp\!\left(\check{i}_k'-\check{r}_k\right)\,,
    \label{eq:losslogSAR2SAR}
\end{align}
with $2\check{\V a}'=\log \V a'^2$, $2\check{\V b}'=\log \V b'^2$, and $\check{\V i}'=\log (\V a'^2+\V b'^2)$ the log-transformed versions of the square of the real and imaginary parts, and the intensity of the \emph{second} noisy realization. Obviously, with half the information available, the same network trained with MERLIN can not perform as well as when trained with the supervised training (i). The degradation in image quality remains limited, however, as seen on figure \ref{fig:resultsMERLINsimus}.

\begin{figure*}[!t]
	\centering
	\includegraphics[width=\textwidth]{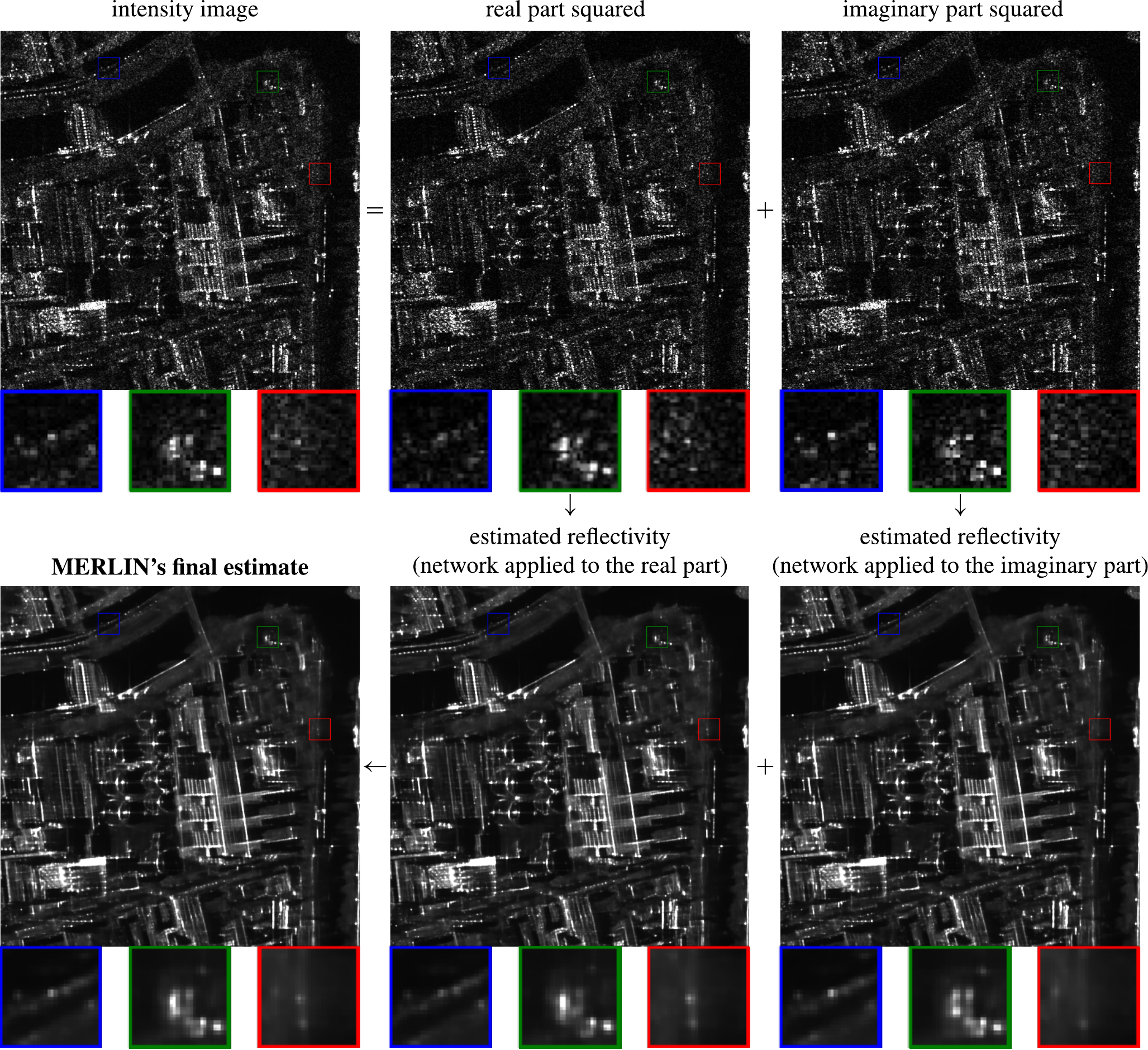}
	\caption{Application of a U-Net trained with MERLIN on a TerraSAR-X image of Berlin, Germany, acquired in \introducedinrevision{high-resolution} spotlight mode. Additional despeckling results on TerraSAR-X images in \introducedinrevision{High Resolution} SpotLight \introducedinrevision{(HS)}  mode can be seen on \url{https://gitlab.telecom-paris.fr/RING/MERLIN}.}
	\label{fig:resultsMERLINreal2}
\end{figure*}
\begin{figure*}[!t]
	\centering
	\includegraphics[width=\textwidth]{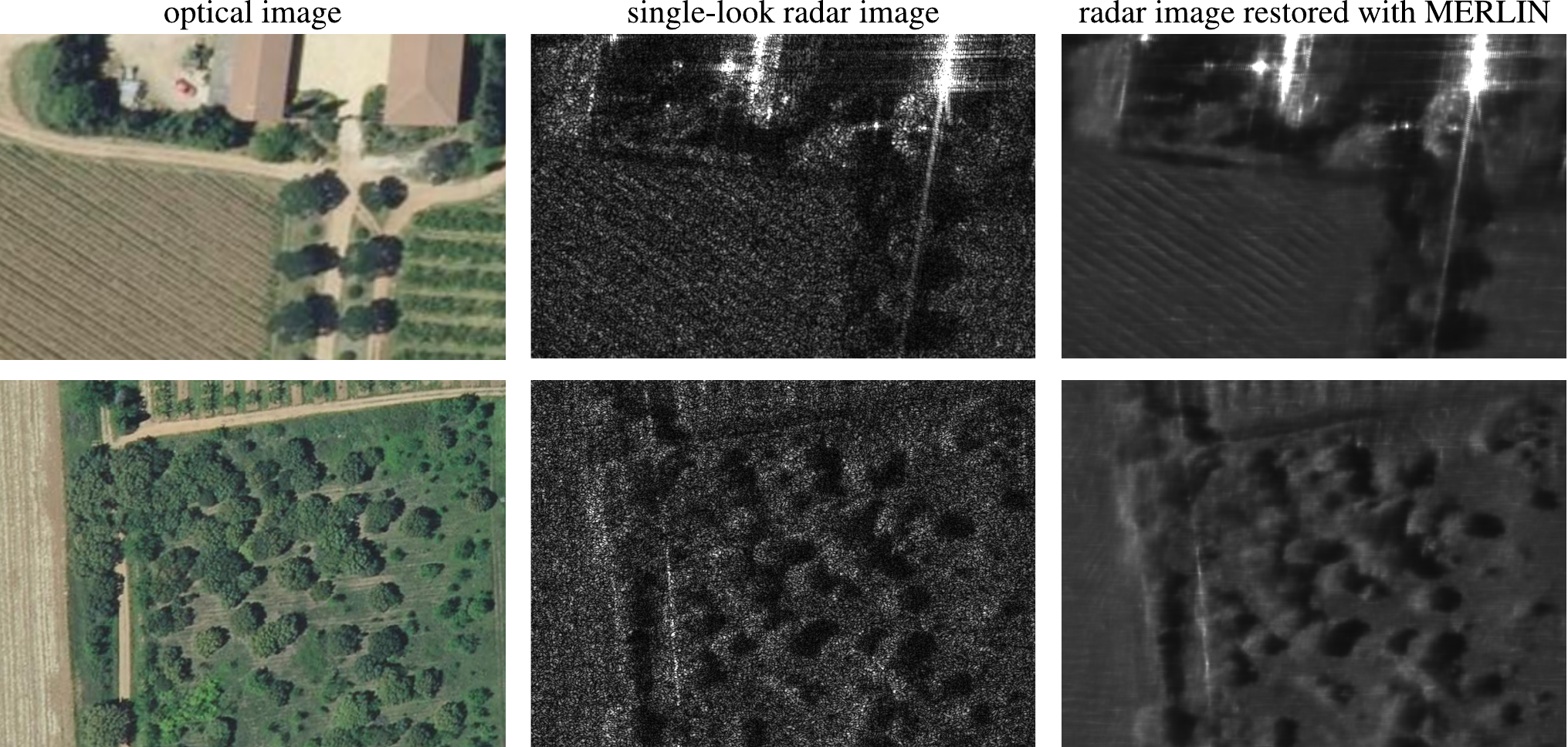}
	\caption{Application of a U-Net trained with MERLIN on a 35cm spatial resolution airborne image on an agricultural area near Nîmes, France, acquired in 2014 by SETHI sensor (\copyright ONERA). The corresponding 20cm resolution optical images (source: Geoportail \copyright IGN), shown on the left, date back from 2018. Additional despeckling results on airborne radar images can be seen on \url{https://gitlab.telecom-paris.fr/RING/MERLIN}.}
	\label{fig:resultsMERLINreal3}
\end{figure*}

Table \ref{table:comparison_psnr} gives PSNR values, expressed on amplitude images $\sqrt{\V r}$, for several despeckling methods. Depending on the image, the U-Net trained with MERLIN performs at least as well or better than methods like SAR-BM3D \cite{parrilli2011nonlocal} or NL-SAR \cite{deledalle2015nl} that are not based on deep neural networks. The performance seems comparable on average to that of SAR-CNN \cite{chierchia2017sar} (this network has been retrained on the same simulated speckle noise images as MERLIN or $\text{SAR2SAR}_\text{A}$, providing the ground truth image to compute the loss function). Numerical values confirm our analysis of figure \ref{fig:resultsMERLINsimus}: when trained with MERLIN, the U-Net produces results that are slightly worse than when real and imaginary parts are processed jointly and the network is trained in a supervised fashion. Compared to the self-supervised method Speckle2Void \cite{molini2020speckle2void} which uses a specific network architecture to obtain a receptive field with a central blind spot, the performance of the U-Net network trained with MERLIN is notably better.
We show in the next paragraph that, when applied to actual SAR images, the gain brought by self-supervision with MERLIN becomes very appealing.

\subsection{Restoration of actual SAR images}
\label{sec:results_real}

\begin{figure*}[!t]
    \centering
    \includegraphics[width=\textwidth]{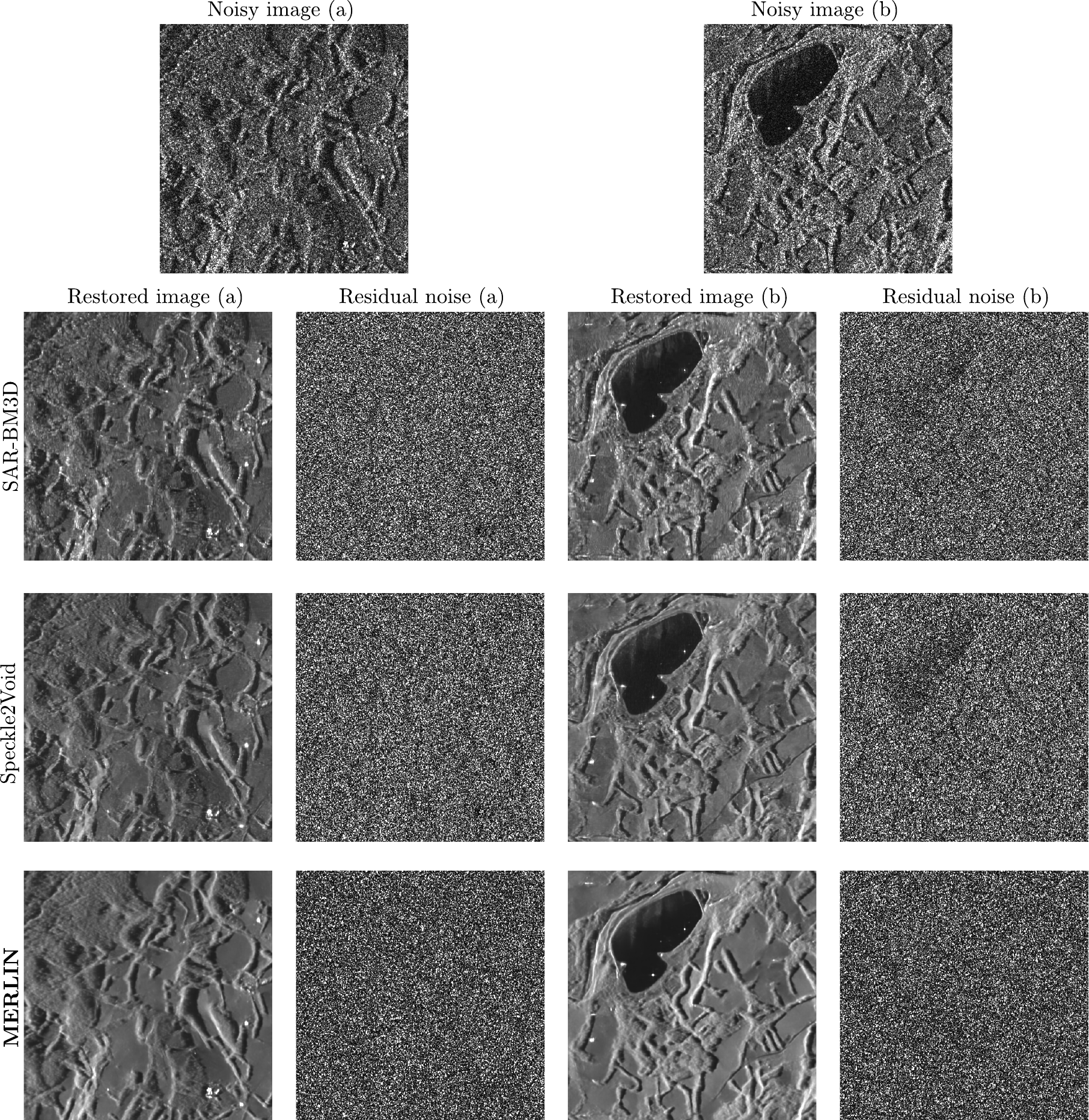}    \caption{Restoration results of three despeckling filters on two TerraSAR-X images near Serre-Ponçon dam, in the French Alps, acquired in stripmap mode. The residual intensity images (\textit{i.e.} the ratio noisy/denoised) is provided to assist in the visual analysis.}
    \label{fig:comparison1}
\end{figure*}

%
%
\begin{figure*}[!t]
    \centering
    \includegraphics[width=\linewidth]{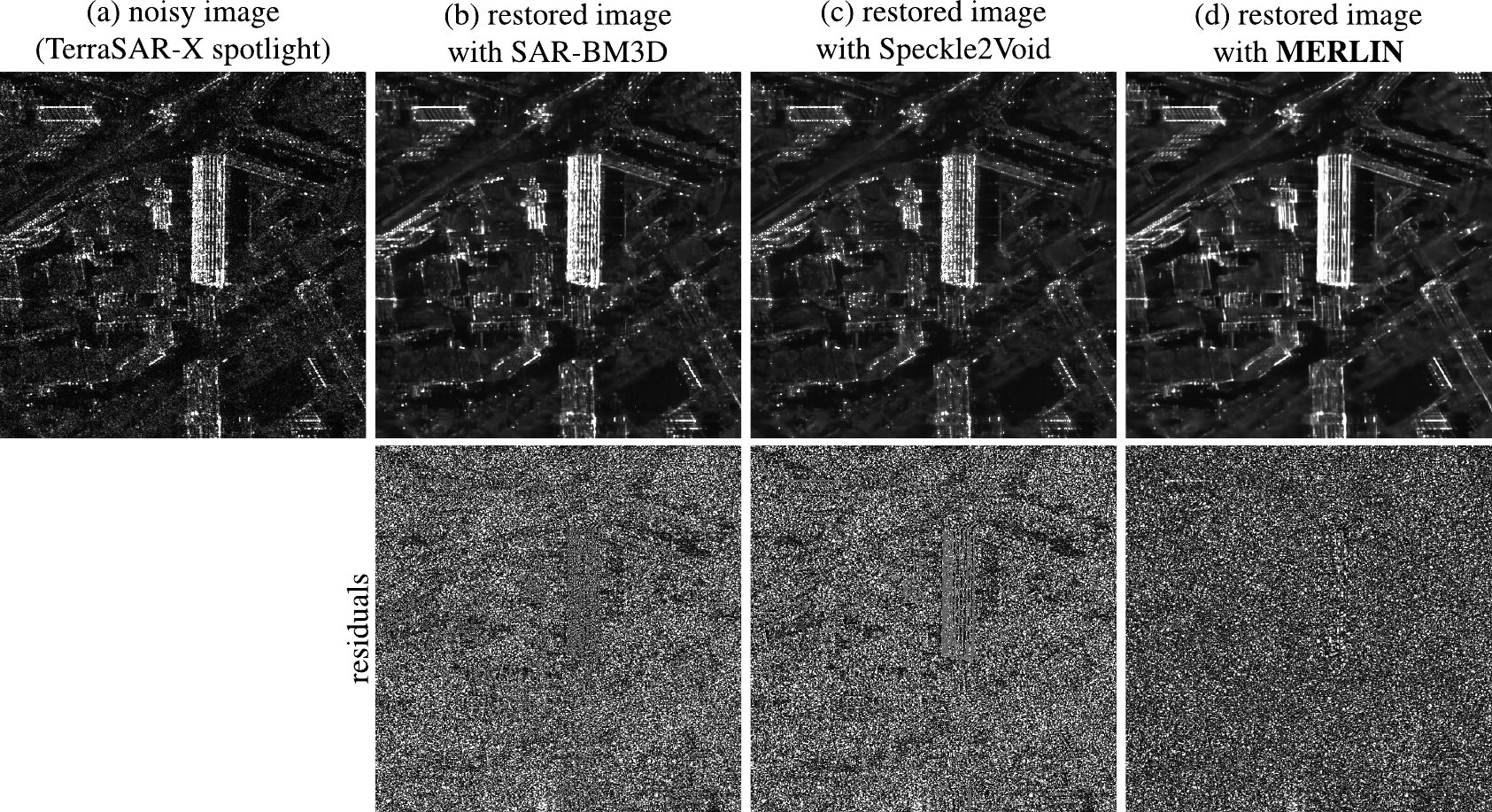}    \caption{Restoration results of three despeckling filters on a TerraSAR-X images of Berlin, in Germany, acquired in \introducedinrevision{high-resolution} spotlight mode. The residual intensity images (\textit{i.e.} the ratio noisy/denoised) is provided to assist in the visual analysis. 
    }
    \label{fig:comparison3}
\end{figure*}
When actual SAR images are considered, it is beneficial to train a network specifically for a given sensor and a particular imaging mode. The SAR transfer function varies from one imaging mode to the other, as well as the spatial resolution and, thus, the structures that can be resolved. In this paragraph, we show results obtained with the same network architecture but different trainings each performed on images of the same type.

Figure \ref{fig:resultsMERLINreal1} shows results obtained on a TerraSAR-X image acquired in stripmap mode over an agricultural area in the French Alps. The first row of the figure illustrates the decomposition of the intensity image into its squared real $\tilde{\V a}^2$ and imaginary $\tilde{\V b}^2$ components. The second row shows the estimations produced by the network trained by MERLIN from each component and the final estimate. A qualitative analysis of the results shows a very good restoration of fine details and textures as well as bright targets.

Figure \ref{fig:resultsMERLINreal2} gives results obtained on a TerraSAR-X image acquired in \introducedinrevision{High Resolution} SpotLight \introducedinrevision{(HS)} mode over an urban area: a small area of the city of Berlin, Germany. The content in this area is very different from the previous region shown in Fig.\ref{fig:resultsMERLINreal1}: there are many bright targets and the images have a higher spatial resolution. Bright targets are preserved while homogeneous areas are smoothed. \introducedinrevision{As highlighted in the zoom-ins of Fig.\ref{fig:resultsMERLINreal2}, point-like scatterers having a deterministic phase appear differently in the real and imaginary part, some of them are even visible only in one of the two parts. Yet, when the intermediate estimations provided by applying the network to the two parts separately are combined together, these targets are correctly restored in the final denoised image.} When many point-like targets are aligned horizontally or vertically (i.e., in the direction of the sidelobes), the network tends to merge the targets into a line.  This phenomenon could possibly be reduced by considering a larger training set and/or a different network architecture.

Figure \ref{fig:resultsMERLINreal3} shows how MERLIN performs in very-high-resolution airborne imaging. The same U-Net network as previously is trained \emph{on a single image} ($9\,130\times10\,000$ pixels) captured with SETHI \cite{baque2017sethi} by the French aerospace laboratory ONERA in 2014. The image has a pixel size of 13cm in range and 19cm in azimuth (the spatial resolution is about 35cm). Two regions of interest are displayed together with optical images at 20cm resolution (orthorectified image by the French geographic institute, IGN). Processing such an image is challenging for a despeckling algorithm because of the spatial correlations of speckle and the strong sidelobes around bright targets. Vegetation seems to be well restored: the stripes visible both in optical and SAR images are preserved and the low-contrasted tree response in the bottom row of Fig.\ref{fig:resultsMERLINreal3} is recovered. Few distortions seem to be applied to bright targets in the top row of Fig.\ref{fig:resultsMERLINreal3}.

Figure \ref{fig:comparison1} \introducedinrevision{and figure \ref{fig:comparison3}} compare the results produced by MERLIN with two other despeckling filters: SAR-BM3D \cite{cozzolino2013fast} and Speckle2Void \cite{molini2020speckle2void}. For each restoration result, the residual image (ratio between the noisy and the denoised image) is also shown. To account for speckle spatial correlations, before applying SAR-BM3D and Speckle2Void, images have been decorrelated. The blind speckle decorrelator proposed by Lapini \textit{et al.} \cite{lapini2013blind} has been used, as suggested by the authors of Speckle2Void \cite{molini2020speckle2void}. Visual inspection of the residual images suggests that some structures have been attenuated by SAR-BM3D, with the edges appearing a bit fuzzy. Moreover, some artifacts arise in homogeneous areas.  

Textures are well recovered by Speckle2Void but a slight bias can be observed in the lake of Fig.\ref{fig:comparison1}.b. \introducedinrevision{Conversely,} the blind-spot structure of the network employed in Speckle2Void makes it hard to recover isolated bright points, as it is difficult to predict their existence from the neighboring pixels. \introducedinrevision{This behavior is exacerbated in dense urban areas, such as that of Figure \ref{fig:comparison3}. 
The images of the residuals appear spatially more stationary with MERLIN than with the other restoration techniques: almost no structure can be identified.
} 
Being robust to speckle spatial correlations and relying on all the pixels in the receptive field of the CNN, MERLIN produces a pleasant result with a good preservation of both geometrical structures and detailed textures, whilst strongly suppressing speckle noise. \introducedinrevision{Point-like scatterers seem to be well restored as well.}

To allow a more extensive evaluations of MERLIN, 
additional restoration results are
provided at \url{https://gitlab.telecom-paris.fr/RING/MERLIN}. 






\section{Discussion}
The statistical model for SAR image formation presented in section \ref{sec:statmod} and figure \ref{fig:statsSAR} served as a basis to derive the loss function used to train networks with MERLIN, Eqs.(\ref{eq:loss}) and (\ref{eq:losslog}). It is based on Goodman's speckle model which is well-established for homogeneous areas imaged at a medium to high resolution. It is known to be less relevant when very high-resolution images are considered, especially in urban areas due to the presence of strong scatterers that dominate the responses within a resolution cell. 
Many alternative statistical models were proposed in the literature \cite{Ovar-15} \cite{Spor-17} \cite{Nico-20} \cite{Dong-21}.
A major difficulty to incorporate such models within MERLIN's loss function is that they depend on additional parameters that would require to be locally set to account for the content of each resolution cell (level of heterogeneity in the cell).
From a pragmatic point of view, the qualitative analysis of the results produced by MERLIN on high-resolution images (Fig.\ref{fig:resultsMERLINreal2}) seems to indicate that the behavior of the network is satisfactory even in the very-high-resolution regime. Performing a more in-depth analysis would probably require using high-quality SAR simulators to provide ground truths for quantitative validation.

Note that, due to the phase modulation applied to perform Terrain Observation with Progressive Scans SAR (TOPSAR) in most of Sentinel-1 acquisition modes (in particular, Interferometric Wide swath, IW), a direct application of MERLIN is \emph{not possible}. It is mandatory that these SLC images be deramped before processing, see \cite{miranda2014definition}. \introducedinrevision{Additional preprocessing also seems necessary to fully decorrelate the real and imaginary parts of these images which will be considered in future works.}

\begin{figure*}[!t]
    \centering
    \includegraphics[width=\linewidth]{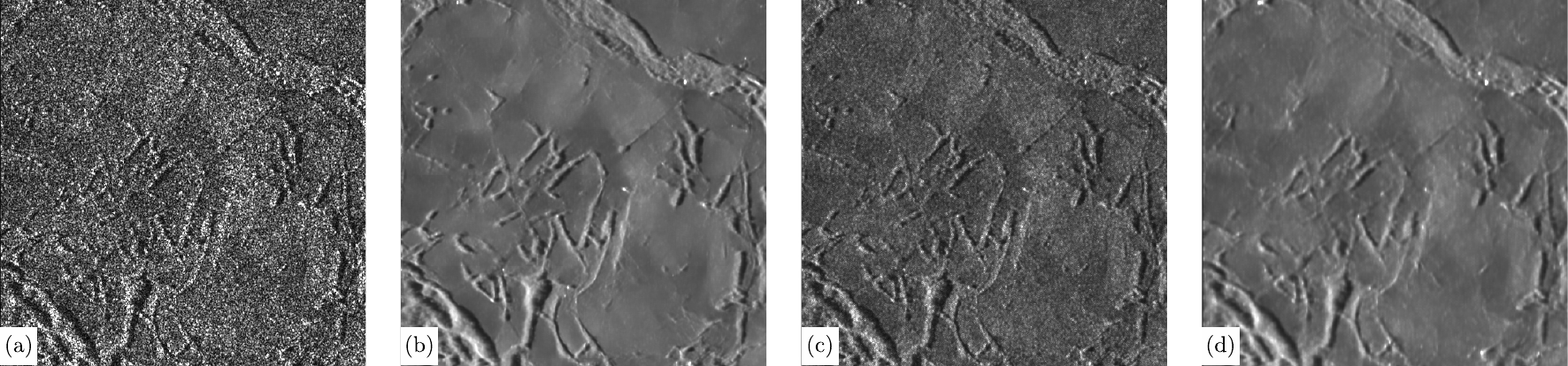}
    \caption{(a) A TerraSAR-X image in stripmap mode. (b) Image restored with MERLIN. (c)-(d) Images restored by a U-Net trained with self-supervision by replacing the real phase with a random phase sampled uniformly in $[-\pi, \pi]$. To produce the result shown in (d) input images are subsampled by a factor of two both at training time and at test time; pixels of the filtered image are then interpolated to recover the original image size.}
    \label{fig:random_phase}
\end{figure*}

One may wonder if MERLIN could possibly work on intensity-only images, by generating fake phase information (a phase could be drawn at each pixel according to a uniform distribution in $[-\pi,\pi]$). This would work perfectly well in the case of an ideal SAR transfer function: the results presented in section \ref{sec:results_simus} would not change if only the intensity was provided to MERLIN and a random phase was generated afterward. When real images are considered, the SAR transfer function is no longer ideal and the statistical distribution of the actual phase differs from a random white field. Figure \ref{fig:random_phase}.c shows the restored images obtained by a network trained on TerraSAR-X Stripmap intensity images, with a phase generated randomly. There are remaining speckle fluctuations and artifacts in the form of a high-frequency texture due to the mismatch between the spatial correlations of the intensity and the phase. Those are absent from the results produced when applying MERLIN on the SLC image (Fig.\ref{fig:random_phase}.b). A possible way to circumvent this problem would consist in subsampling the images to decrease the speckle spatial correlations, both at training and at testing time, and draw random phases. This leads to images free of correlation artifacts, see Fig.\ref{fig:random_phase}.d, at the cost of a degradation of the highest-frequency details (such as thin lines). In conclusion, it is preferable to use SLC images when available to directly apply MERLIN. If only the intensity is available, then the speckle should first be spatially whitened (by inverting the SAR transfer function \cite{lapini2013blind,abergel2018subpixellic} or by subsampling \cite{dalsasso2020handle}) before applying MERLIN on the pseudo-SLC image obtained by drawing random phases.

Another point worth discussion is the inversion of the SAR transfer function.
The loss functions in Eqs.(\ref{eq:loss}) and (\ref{eq:losslog}) were derived from the \emph{marginal distributions}. Starting from the full distribution, a different loss function would be obtained:
\begin{align}
    \mathcal{L}_\text{full}({\V r},\tilde{\V b})&=
    \sum_k \left(\frac{1}{2}\log  r_k\right)
    +\tilde{\V b}\transpose\!\!\left[\M H\text{diag}(\V r)\M H\transpose\right]^{-1}\!\tilde{\V b}\,,
    \label{eq:lossfull}
\end{align}
which boils down to Eq.(\ref{eq:loss}) when $\M H=\M I$.
The loss function of Eq.(\ref{eq:lossfull}) is much more costly to evaluate since $\M H$ corresponds to a Toeplitz-block Toeplitz matrix (a 2D convolution in the direct domain, or a product in the Fourier domain). The inversion $\left[\M H\text{diag}(\V r)\M H\transpose\right]^{-1}$ can not be derived in closed form, which makes the training much more costly (several iterations of a conjugate gradients algorithm would typically be necessary for each evaluation of the loss function). Moreover, some form of regularization on $\V r$ would be necessary since the data $\tilde{\V a}$ and $\tilde{\V b}$ are not sufficient to constrain $\V r$ beyond the cutoff frequency of the SAR system.

As illustrated in Fig.\ref{fig:resultsMERLINsimus} and commented in section \ref{sec:results_simus}, the self-supervision with MERLIN comes at a cost: the real and imaginary components are processed separately. This limits the performance compared to joint processing of both components since the network is forced to handle images with a worse SNR (by a factor $1/\sqrt{2}$). 
This is partially compensated when the restorations computed separately on each component are finally combined. We think that this drawback is amply compensated by the good adaptation of the network to the SAR transfer function and the ability with MERLIN to use very large training sets (possibly, entire archives from a sensor).

Finally, compared to other self-supervision approaches, MERLIN imposes no limitation to the architecture of the network and does not assume a spatially decorrelated speckle. Together with the possibility to straightforwardly include huge archives of images in the training set, this opens new possibilities to consider highly expressive network architectures (e.g., very deep) for despeckling.

\section{Conclusion}
We have shown that single-look complex images offer an ideal framework to self-train despeckling networks \FT{using the real and imaginary parts of the data}. The proposed generic training approach, called MERLIN, imposes no constrain on the architecture of the network. It completely suppresses the hassle of building training sets with reference images. We believe that this will dramatically change the way deep despeckling networks are used: networks specific to a sensor/acquisition mode can be easily trained, reaching a higher performance than general-purpose networks. Relieved from the worry of building training sets, future work can focus on designing clever network architectures. With MERLIN, very large scale training using entire archives of SAR images produced with a specific sensor mode can be contemplated, which could potentially lead to unprecedented despeckling performances.

\section*{Acknowledgments}
The airborne SAR images processed in this paper were provided by ONERA, the French aerospace lab, within the project ALYS ANR-15-ASTR-0002 funded by the DGA (Direction Générale à l'Armement) and the ANR (Agence Nationale de la Recherche). Some of the TerraSAR-X images have been provided by the German Space Agency DLR for the project DLR-MTH0232.
\introducedinrevision{The authors are grateful to the Reviewers for their comments that helped to improve the paper, in particular for pointing out that it is essential to require the radar response $\M H$ to be real-valued (proper 0-doppler centering).}

\appendices
\introducedinrevision{
\section{Condition on the SAR response to obtain independent real and imaginary parts}
\label{sec:appendixindep}
MERLIN self-supervised training procedure is based on the statistical independence of $\tilde{\V a}$ and $\tilde{\V b}$, the real and imaginary parts of the single-look complex SAR image. Under Goodman's fully-developed speckle model, the real and imaginary parts $\V a$ and $\V b$ obtained with an ideal SAR system are i.i.d. under a Gaussian distribution $\mathcal{N}\left(\V 0,\tfrac{1}{2}\text{diag}(\V r)\right)$, i.e., $\V z\equiv \V a+j\V b$ follows a circular complex Gaussian distribution with covariance $\tfrac{1}{2}\text{diag}(\V r)$. A radar system modeled by the linear (possibly complex-valued) spatial-domain operator $\mathbf{H}$ produces correlated samples $\tilde{\V z}=\mathbf{H}\V z$ distributed according to a circular complex Gaussian distribution with covariance matrix $\mathbf C=\mathbf{H}\text{diag}(\V r)\mathbf{H}\transpose$ (where $\cdot\transpose$ denotes the conjugate transpose). The real and imaginary parts $\tilde{\V a}$ and $\tilde{\V b}$ are jointly distributed according to a Gaussian distribution:
\begin{align}
    \begin{pmatrix}
    \tilde{\V a}\\
    \tilde{\V b}
    \end{pmatrix}\sim
    \mathcal{N}\left(\begin{pmatrix}\V 0\\
    \V 0\end{pmatrix}
    ,\M \Sigma\right)\,,
\end{align}
with $\M \Sigma$ the covariance matrix defined by blocks:
\begin{align}
    \M \Sigma&=\tfrac{1}{2}
    \begin{pmatrix}
    \M M&-\M N\\
    \M N&\M M
    \end{pmatrix}
    \begin{pmatrix}
    \text{diag}(\V r)&\M 0\\
    \M 0&\text{diag}(\V r)
    \end{pmatrix}
    \begin{pmatrix}
    \M M\transpose&\M N\transpose\\
    -\M N\transpose&\M M\transpose
    \end{pmatrix}\,,
    \label{eq:sigmacorrel}
\end{align}
and $\M H=\M M+j\M N$. The expansion of equation (\ref{eq:sigmacorrel}) leads to the following expressions for the 4 blocks:
\begin{align}
\begin{cases}
    2\M\Sigma_{1,1}=
    \M M\text{diag}(\V r)\M M\transpose+
    \M N\text{diag}(\V r)\M N\transpose\\
    2\M\Sigma_{1,2}=\M M\text{diag}(\V r)\M N\transpose-
    \M N\text{diag}(\V r)\M M\transpose\\
    2\M\Sigma_{2,1}=\M N\text{diag}(\V r)\M M\transpose-
    \M M\text{diag}(\V r)\M N\transpose\\
    2\M\Sigma_{2,2}=
    \M M\text{diag}(\V r)\M M\transpose+
    \M N\text{diag}(\V r)\M N\transpose.\\
    \end{cases}
\end{align}
This shows that $\tilde{\V a}$ and $\tilde{\V b}$ are independent if and only if 
\begin{align}
\M M\text{diag}(\V r)\M N\transpose=\M N\text{diag}(\V r)\M M\transpose\,,
\label{eq:conddecorr}
\end{align}
for all reflectivities $\V r$ (so that $\M\Sigma_{1,2}=\M\Sigma_{2,1}=\M 0$). By writing $\M M=(\V m_1\cdots \V m_K)$, with $\V m_k$ the $k$-th column of $\M M$, $\M N=(\V n_1\cdots \V n_K)$, and by choosing $\V r=\V e_i$ (the $i$-th elementary vector, i.e., $r_k=0$ if $k\neq i$ and $r_i=1$), equation (\ref{eq:conddecorr}) gives:
\begin{align}
\V m_i \V n_i\transpose=\V n_i \V m_i\transpose\,.
\label{eq:condrankone}
\end{align}
Matrix $\V m_i \V n_i\transpose$ is thus symmetric and at most of rank one. Vectors $\V m_i$ and $\V n_i$ are thus collinear: there exist a couple of scalars $(\lambda_i,\tau_i)$ and a vector $\V q_i\in\mathbb{R}^K$ such that $\V m_i=\lambda_i\V q_i$ and $\V n_i=\tau_i\V q_i$ (setting $\lambda_i$ or $\tau_i$ to 0 leads to a rank 0 matrix).
Therefore, condition (\ref{eq:conddecorr}) implies that matrices $\M M$ and $\M N$ take the form: 
\begin{align}
\M M=\M Q\text{diag}(\V\lambda)\qquad\text{and}\qquad\M N=\M Q\text{diag}(\V\tau)\,,
\label{eq:condmatgene}
\end{align}
with $\V\lambda$ and $\V\tau$ any vector of $\mathbb{R}^K$. It can be easily checked that this necessary condition is sufficient for equation (\ref{eq:conddecorr}) to hold. Real and imaginary components $\tilde{\V a}$ and $\tilde{\V b}$ are thus statistically independent if and only if the SAR system response $\M H$ can be written under the form:
\begin{align}
\M H&=\M Q\text{diag}(\V\lambda+j\V\tau)\\
&=\M Q'\text{diag}\left(\exp(j\V\varphi)\right)\,,
\label{eq:condH}
\end{align}
where $\M Q$ and $\M Q'$ are real-valued linear operators and $\exp(j\V\varphi)$ is a $K$-dimensional phase vector (i.e., a vector of $K$ complex values, each with unit magnitude). Equation (\ref{eq:condH}) shows that the SAR response must take the form of a per-pixel phase shift followed by a real-valued linear operator.
\\
\indent If the SAR system has a shift-invariant response, then all $\varphi_i$ are equal and $\M H=\exp(j\varphi)\M Q'$ (where $\varphi$ is a scalar): a constant phase shift is applied to the whole image. Linear operators $\M H$ and $\M Q'$ can then be diagonalized by the 2D discrete Fourier transform: $\M H=\M F\transpose \text{diag}(\bar{\V h})\M F$, with $\bar{\V h}$ the frequency response of the SAR system, and $\M Q'=\M F\transpose \text{diag}(\bar{\V q}')\M F$ (where $\bar{\V q}'$ is the frequency response of $\M Q'$). Since $\M H=\exp(j\varphi)\M Q'$, $|\bar{\V h}|=|\exp(j\varphi)\bar{\V q}'|=|\bar{\V q}'|$. Given that $\M Q'$ is real-valued, $|\bar{\V q}'|$ is even and the gain of the SAR system $|\bar{\V h}|$ must also be even in order for the system to produce independent real and imaginary parts $\tilde{\V a}$ and $\tilde{\V b}$.
}

\introducedinrevision{
	\section{Pre-processing of image patches to obtain independent real and imaginary parts}
	\label{sec:appendixpreprocessing}
	Due to a non-zero Doppler centroid or an acquisition mode with a time-varying squint angle, the frequency response of the SAR system may be asymmetrical, leading to correlated real and imaginary parts in the SAR images. This would prevent the application of MERLIN. To avoid this problem, we preprocess each patch to recenter its spectrum. The U-Net employed in this paper takes as input patches with $256\times 256$ pixels. These patches are preprocessed as follows.\\[1ex]
	%
	\indent To correct for a non-zero Doppler at the scale of the patch, we compute the azimuth and range 1D profiles by averaging the magnitude of the 2D Fourier transform of the patch along the range and azimuth directions, respectively. Let $\V p$ denote one such profile. We look for the shift $\delta$ such that the translated profile $T_{\delta}\{\V p\}$ and its symmetric $\mathcal{S}\{T_{\delta}\{\V p\}\}$ superimpose at best: $\widehat\delta=\mathop{\text{arg min}}_{\delta} ||T_{\delta}\{\V p\}-\mathcal{S}\{T_{\delta}\{\V p\}\}||_2^2$. Since the translations of the spectrum that we consider are circular, $||T_{\delta}\{\V p\}||_2^2$ and $||\mathcal{S}\{T_{\delta}\{\V p\}\}||_2^2$ are constant for all values of $\delta$. It is then equivalent to estimate $\delta$ by maximizing the correlation: $\widehat\delta=\mathop{\text{arg max}}_{\delta} T_{\delta}\{\V p\}\transpose\mathcal{S}\{T_{\delta}\{\V p\}\}$. In order to evaluate efficiently this cross-correlation for all integer shifts $\delta$, we consider the equivalent formulation $\widehat\delta=\mathop{\text{arg max}}_{\delta} T_{2\delta}\{\V p\}\transpose\mathcal{S}\{\V p\}=\mathop{\text{arg max}}_{\delta} \left[\V p\star\mathcal{S}\{\V p\}\right](2\delta)$, where $\star$ denotes the discrete correlation (computed in Fourier domain with fast Fourier transforms).\\[1ex]
	\indent The airborne image shown in Fig.\ref{fig:resultsMERLINreal3} has a spectrum with an asymmetrical support. After centering the spectrum, we built a mask corresponding to all frequencies $(\nu_x,\nu_y)$ such that their symmetric counterparts $(-\nu_x,\nu_y)$, $(\nu_x,-\nu_y)$, and $(-\nu_x,-\nu_y)$ are all within the bandwidth of the SAR system (i.e., non-zero). We then cut all frequencies off beyond our symmetrical mask to ensure the symmetry of the spectrum.
}

\bibliographystyle{IEEEtran}
\bibliography{ref}

\begin{thebibliography}{10}
\providecommand{\url}[1]{#1}
\csname url@samestyle\endcsname
\providecommand{\newblock}{\relax}
\providecommand{\bibinfo}[2]{#2}
\providecommand{\BIBentrySTDinterwordspacing}{\spaceskip=0pt\relax}
\providecommand{\BIBentryALTinterwordstretchfactor}{4}
\providecommand{\BIBentryALTinterwordspacing}{\spaceskip=\fontdimen2\font plus
\BIBentryALTinterwordstretchfactor\fontdimen3\font minus
  \fontdimen4\font\relax}
\providecommand{\BIBforeignlanguage}[2]{{%
\expandafter\ifx\csname l@#1\endcsname\relax
\typeout{** WARNING: IEEEtran.bst: No hyphenation pattern has been}%
\typeout{** loaded for the language `#1'. Using the pattern for}%
\typeout{** the default language instead.}%
\else
\language=\csname l@#1\endcsname
\fi
#2}}
\providecommand{\BIBdecl}{\relax}
\BIBdecl

\bibitem{lee1983digital}
J.~Lee, ``Digital image smoothing and the sigma filter,'' \emph{Computer
  vision, graphics, and image processing}, vol.~24, no.~2, pp. 255--269, 1983.

\bibitem{fracastoro2020deep}
G.~Fracastoro, E.~Magli, G.~Poggi, G.~Scarpa, D.~Valsesia, and L.~Verdoliva,
  ``{Deep learning methods for SAR image despeckling: trends and
  perspectives},'' \emph{arXiv preprint arXiv:2012.05508}, 2020.

\bibitem{zhu2021deep}
X.~Zhu, S.~Montazeri, M.~Ali, Y.~Hua, Y.~Wang, L.~Mou, Y.~Shi, F.~Xu, and
  R.~Bamler, ``{Deep learning meets SAR: concepts, models, pitfalls, and
  perspectives},'' \emph{IEEE Geoscience and Remote Sensing Magazine (GRSM)},
  2021.

\bibitem{denis2021review}
L.~Denis, E.~Dalsasso, and F.~Tupin, ``{A review of deep-learning techniques
  for SAR image restoration},'' in \emph{IEEE International Geoscience and
  Remote Sensing Symposium (IGARSS)}.\hskip 1em plus 0.5em minus 0.4em\relax
  IEEE, 2021.

\bibitem{vasile2006intensity}
G.~Vasile, E.~Trouv{\'e}, J.~Lee, and V.~Buzuloiu, ``{Intensity-driven
  adaptive-neighborhood technique for polarimetric and interferometric SAR
  parameters estimation},'' \emph{IEEE Trans. Geos. Remote Sens.}, vol.~44,
  no.~6, pp. 1609--1621, 2006.

\bibitem{lopes1990adaptive}
A.~Lopes, R.~Touzi, and E.~Nezry, ``Adaptive speckle filters and scene
  heterogeneity,'' \emph{IEEE Trans. Geos. Remote Sens.}, vol.~28, no.~6, pp.
  992--1000, 1990.

\bibitem{aubert2008variational}
G.~Aubert and J.-F. Aujol, ``A variational approach to removing multiplicative
  noise,'' \emph{SIAM journal on applied mathematics}, vol.~68, no.~4, pp.
  925--946, 2008.

\bibitem{denis2009sar}
L.~Denis, F.~Tupin, J.~Darbon, and M.~Sigelle, ``{SAR image regularization with
  fast approximate discrete minimization},'' \emph{IEEE Trans. Image Proc.},
  vol.~18, no.~7, pp. 1588--1600, 2009.

\bibitem{bioucas2010multiplicative}
J.~M. Bioucas-Dias and M.~A. Figueiredo, ``Multiplicative noise removal using
  variable splitting and constrained optimization,'' \emph{IEEE Trans. Image
  Proc.}, vol.~19, no.~7, pp. 1720--1730, 2010.

\bibitem{steidl2010removing}
G.~Steidl and T.~Teuber, ``{Removing multiplicative noise by Douglas-Rachford
  splitting methods},'' \emph{Journal of Mathematical Imaging and Vision},
  vol.~36, no.~2, pp. 168--184, 2010.

\bibitem{aujol2003image}
J.-F. Aujol, G.~Aubert, L.~Blanc-F{\'e}raud, and A.~Chambolle, ``{Image
  decomposition application to SAR images},'' in \emph{International Conference
  on Scale-Space Theories in Computer Vision}.\hskip 1em plus 0.5em minus
  0.4em\relax Springer, 2003, pp. 297--312.

\bibitem{lobry2016multitemporal}
S.~Lobry, L.~Denis, and F.~Tupin, ``{Multitemporal SAR image decomposition into
  strong scatterers, background, and speckle},'' \emph{IEEE Jour. Sel. Top.
  App. Earth Obs. Remote Sens.}, vol.~9, no.~8, pp. 3419--3429, 2016.

\bibitem{xie2002sar}
H.~Xie, L.~E. Pierce, and F.~T. Ulaby, ``{SAR speckle reduction using wavelet
  denoising and Markov random field modeling},'' \emph{IEEE Trans. Geos. Remote
  Sens.}, vol.~40, no.~10, pp. 2196--2212, 2002.

\bibitem{durand2009multiplicative}
S.~Durand, J.~Fadili, and M.~Nikolova, ``Multiplicative noise cleaning via a
  variational method involving curvelet coefficients,'' in \emph{International
  Conference on Scale Space and Variational Methods in Computer Vision}.\hskip
  1em plus 0.5em minus 0.4em\relax Springer, 2009, pp. 282--294.

\bibitem{NLM}
A.~Buades, B.~Coll, and J.~M. Morel, ``A review of image denoising algorithms,
  with a new one,'' \emph{Simul}, vol.~4, pp. 490--530, 2005.

\bibitem{BM3Dwiener}
K.~Dabov, A.~Foi, V.~Katkovnik, and K.~Egiazarian, ``{Image Denoising by Sparse
  3-D Transform-Domain Collaborative Filtering},'' \emph{IEEE Trans. Imag.
  Proc.}, vol.~16, no.~8, pp. 2080--2095, {A}gu.

\bibitem{deledalle2014exploiting}
C.-A. Deledalle, L.~Denis, G.~Poggi, F.~Tupin, and L.~Verdoliva, ``{Exploiting
  patch similarity for SAR image processing: the nonlocal paradigm},''
  \emph{IEEE Sig. Proc. Mag.}, vol.~31, no.~4, pp. 69--78, 2014.

\bibitem{tupin2019ten}
F.~Tupin, L.~Denis, C.-A. Deledalle, and G.~Ferraioli, ``{Ten Years of
  Patch-Based Approaches for SAR Imaging: A Review},'' in \emph{IGARSS
  2019-2019 IEEE International Geoscience and Remote Sensing Symposium}.\hskip
  1em plus 0.5em minus 0.4em\relax IEEE, 2019, pp. 5105--5108.

\bibitem{deledalle2009iterative}
C.-A. Deledalle, L.~Denis, and F.~Tupin, ``Iterative weighted maximum
  likelihood denoising with probabilistic patch-based weights,'' \emph{IEEE
  Transactions on Image Processing}, vol.~18, no.~12, pp. 2661--2672, 2009.

\bibitem{parrilli2011nonlocal}
S.~Parrilli, M.~Poderico, C.~V. Angelino, and L.~Verdoliva, ``{A nonlocal SAR
  image denoising algorithm based on LLMMSE wavelet shrinkage},'' \emph{IEEE
  Transactions on Geoscience and Remote Sensing}, vol.~50, no.~2, pp. 606--616,
  2011.

\bibitem{cozzolino2013fast}
D.~Cozzolino, S.~Parrilli, G.~Scarpa, G.~Poggi, and L.~Verdoliva, ``{Fast
  adaptive nonlocal SAR despeckling},'' \emph{IEEE Geoscience and Remote
  Sensing Letters}, vol.~11, no.~2, pp. 524--528, 2013.

\bibitem{deledalle2010nl}
C.-A. Deledalle, L.~Denis, and F.~Tupin, ``{NL-InSAR: Nonlocal interferogram
  estimation},'' \emph{IEEE Transactions on Geoscience and Remote Sensing},
  vol.~49, no.~4, pp. 1441--1452, 2010.

\bibitem{chen2010nonlocal}
J.~Chen, Y.~Chen, W.~An, Y.~Cui, and J.~Yang, ``{Nonlocal filtering for
  polarimetric SAR data: A pretest approach},'' \emph{IEEE Transactions on
  Geoscience and Remote Sensing}, vol.~49, no.~5, pp. 1744--1754, 2010.

\bibitem{torres2014speckle}
L.~Torres, S.~J. Sant'Anna, C.~da~Costa~Freitas, and A.~C. Frery, ``{Speckle
  reduction in polarimetric SAR imagery with stochastic distances and nonlocal
  means},'' \emph{Pattern Recognition}, vol.~47, no.~1, pp. 141--157, 2014.

\bibitem{deledalle2014nl}
C.-A. Deledalle, L.~Denis, F.~Tupin, A.~Reigber, and M.~J{\"a}ger, ``{NL-SAR: A
  unified nonlocal framework for resolution-preserving (Pol)(In) SAR
  denoising},'' \emph{IEEE Transactions on Geoscience and Remote Sensing},
  vol.~53, no.~4, pp. 2021--2038, 2014.

\bibitem{wang2017sar}
P.~Wang, H.~Zhang, and V.~M. Patel, ``{SAR} image despeckling using a
  convolutional neural network,'' \emph{IEEE Sig. Proces. Let.}, vol.~24,
  no.~12, pp. 1763--1767, 2017.

\bibitem{wang2017generative}
------, ``Generative adversarial network-based restoration of speckled {SAR}
  images,'' in \emph{2017 IEEE CAMSAP}.\hskip 1em plus 0.5em minus 0.4em\relax
  IEEE, 2017, pp. 1--5.

\bibitem{zhang2018learning}
Q.~Zhang, Q.~Yuan, J.~Li, Z.~Yang, and X.~Ma, ``Learning a dilated residual
  network for {SAR} image despeckling,'' \emph{Remote Sens.}, vol.~10, no.~2,
  p. 196, 2018.

\bibitem{lattari2019deep}
F.~Lattari, B.~Gonzalez~Leon, F.~Asaro, A.~Rucci, C.~Prati, and M.~Matteucci,
  ``Deep learning for {SAR} image despeckling,'' \emph{Remote Sens.}, vol.~11,
  no.~13, p. 1532, 2019.

\bibitem{dalsasso2020sar}
E.~Dalsasso, X.~Yang, L.~Denis, F.~Tupin, and W.~Yang, ``{SAR} {I}mage
  {D}especkling by {D}eep {N}eural {N}etworks: from a pre-trained model to an
  end-to-end training strategy,'' \emph{Remote Sens.}, vol.~12, no.~16, p.
  2636, 2020.

\bibitem{lapini2013blind}
A.~Lapini, T.~Bianchi, F.~Argenti, and L.~Alparone, ``Blind speckle
  decorrelation for {SAR} image despeckling,'' \emph{IEEE Trans. Geos. Remote
  Sens.}, vol.~52, no.~2, pp. 1044--1058, 2013.

\bibitem{abergel2018subpixellic}
R.~Abergel, L.~Denis, S.~Ladjal, and F.~Tupin, ``{Subpixellic methods for
  sidelobes suppression and strong targets extraction in single look complex
  SAR images},'' \emph{IEEE Journal of Selected Topics in Applied Earth
  Observations and Remote Sensing}, vol.~11, no.~3, pp. 759--776, 2018.

\bibitem{dalsasso2020handle}
E.~Dalsasso, L.~Denis, and F.~Tupin, ``{How to handle spatial correlations in
  SAR despeckling? Resampling strategies and deep learning approaches},'' in
  \emph{{13th European Conference on Synthetic Aperture Radar (EUSAR)}}.\hskip
  1em plus 0.5em minus 0.4em\relax {VDE ITG}, 2021, pp. 1233--1238.

\bibitem{chierchia2017sar}
G.~Chierchia, D.~Cozzolino, G.~Poggi, and L.~Verdoliva, ``{SAR image
  despeckling through convolutional neural networks},'' in \emph{2017 IEEE
  International Geoscience and Remote Sensing Symposium (IGARSS)}.\hskip 1em
  plus 0.5em minus 0.4em\relax IEEE, 2017, pp. 5438--5441.

\bibitem{cozzolino2020nonlocal}
D.~Cozzolino, L.~Verdoliva, G.~Scarpa, and G.~Poggi, ``Nonlocal {CNN} {SAR}
  {I}mage {D}especkling,'' \emph{Remote Sens.}, vol.~12, no.~6, p. 1006, 2020.

\bibitem{dalsasso2021sar2sar}
E.~Dalsasso, L.~Denis, and F.~Tupin, ``{SAR2SAR: a semi-supervised despeckling
  algorithm for SAR images},'' \emph{IEEE Journal of Selected Topics in Applied
  Earth Observations and Remote Sensing}, vol.~14, pp. 4321--4329, 2021.

\bibitem{9324183}
A.~B. Molini, D.~Valsesia, G.~Fracastoro, and E.~Magli, ``{Towards Deep
  Unsupervised Sar Despeckling with Blind-Spot Convolutional Neural
  Networks},'' in \emph{IGARSS 2020 - 2020 IEEE International Geoscience and
  Remote Sensing Symposium}, 2020, pp. 2507--2510.

\bibitem{molini2020speckle2void}
------, ``Speckle2void: Deep self-supervised sar despeckling with blind-spot
  convolutional neural networks,'' \emph{IEEE Transactions on Geoscience and
  Remote Sensing}, 2021.

\bibitem{laine2019high}
S.~Laine, T.~Karras, J.~Lehtinen, and T.~Aila, ``High-quality self-supervised
  deep image denoising,'' in \emph{Advances in Neural Information Processing
  Systems}, 2019, pp. 6970--6980.

\bibitem{lee2020noise2kernel}
K.~Lee and W.-K. Jeong, ``{Noise2Kernel: Adaptive Self-Supervised Blind
  Denoising using a Dilated Convolutional Kernel Architecture},'' \emph{arXiv
  preprint arXiv:2012.03623}, 2020.

\bibitem{xie2020noise2same}
Y.~Xie, Z.~Wang, and S.~Ji, ``{Noise2Same: Optimizing A Self-Supervised Bound
  for Image Denoising},'' \emph{arXiv preprint arXiv:2010.11971}, 2020.

\bibitem{lehtinen2018noise2noise}
J.~Lehtinen, J.~Munkberg, J.~Hasselgren, S.~Laine, T.~Karras, M.~Aittala, and
  T.~Aila, ``{Noise2Noise: Learning Image Restoration without Clean Data},'' in
  \emph{International Conference on Machine Learning}.\hskip 1em plus 0.5em
  minus 0.4em\relax PMLR, 2018, pp. 2965--2974.

\bibitem{goodman2007speckle}
J.~W. Goodman, \emph{Speckle phenomena in optics: theory and
  applications}.\hskip 1em plus 0.5em minus 0.4em\relax Roberts and Company
  Publishers, 2007.

\bibitem{ronneberger2015u}
O.~Ronneberger, P.~Fischer, and T.~Brox, ``{U-Net: Convolutional networks for
  biomedical image segmentation},'' in \emph{International Conference on
  Medical image computing and computer-assisted intervention}.\hskip 1em plus
  0.5em minus 0.4em\relax Springer, 2015, pp. 234--241.

\bibitem{zhang2019gradient}
J.~Zhang, T.~He, S.~Sra, and A.~Jadbabaie, ``Why gradient clipping accelerates
  training: A theoretical justification for adaptivity,'' \emph{arXiv preprint
  arXiv:1905.11881}, 2019.

\bibitem{deledalle2015nl}
C.-A. Deledalle, L.~Denis, F.~Tupin, A.~Reigber, and M.~J{\"a}ger, ``{NL-SAR}:
  A unified nonlocal framework for resolution-preserving {(Pol)(In) SAR}
  denoising,'' \emph{IEEE TGRS}, vol.~53, no.~4, pp. 2021--2038, 2015.

\bibitem{baque2017sethi}
R.~Baqu{\'e}, O.~R. du~Plessis, N.~Castet, P.~Fromage, J.~Martinot-Lagarde,
  J.-F. Nouvel, H.~Oriot, S.~Angelliaume, F.~Brigui, H.~Cantalloube
  \emph{et~al.}, ``{SETHI/RAMSES-NG: New performances of the flexible
  multi-spectral airborne remote sensing research platform},'' in \emph{2017
  European Radar Conference (EURAD)}.\hskip 1em plus 0.5em minus 0.4em\relax
  IEEE, 2017, pp. 191--194.

\bibitem{Ovar-15}
J.-P. Ovarlez, F.~Pascal, and P.~Forster, ``{Covariance Matrix Estimation in
  SIRV and Elliptical Processes and Their Applications in Radar Detection},''
  in \emph{Modern Radar Detection}, 2015, pp. 295--332.

\bibitem{Spor-17}
H.~Sportouche, J.-M. Nicolas, and F.~Tupin, ``{Mimic Capacity Of Fisher And
  Generalized Gamma Distributions For High Resolution SAR Image Statistical
  Modeling},'' \emph{IEEE Jour. Sel. Top. App. Earth Obs. Remote Sens.},
  vol.~10, no.~12, pp. 5724--5735, 2017.

\bibitem{Nico-20}
J.-M. Nicolas and F.~Tupin, ``{A New Parameterization for the Rician
  Distribution},'' \emph{IEEE Geoscience and Remote Sensing Letters}, vol.~17,
  no.~11, 2020.

\bibitem{Dong-21}
D.-X. Yue, F.~Xu, A.~C. Frery, and Y.-Q. Jin, ``{Synthetic Aperture Radar Image
  Statistical Modeling: Part One - Single-Pixel Statistical Models},''
  \emph{IEEE Geoscience and Remote Sensing Magazine}, vol.~9, no.~1, 2021.

\bibitem{miranda2014definition}
N.~Miranda, ``{Definition of the TOPS SLC deramping function for products
  generated by the S-1 IPF},'' \emph{Eur. Space Agency, Paris, France, Tech.
  Rep}, 2014.

\end{thebibliography}

\end{document}